%% file: root.tex
\documentclass[letterpaper, 10 pt, conference]{ieeeconf}  

\IEEEoverridecommandlockouts                              

\usepackage[sortcites,backend=bibtex,citestyle=numeric,bibstyle=ieee,sorting=none,natbib=true,maxcitenames=1,mincitenames=1]{biblatex}

\usepackage{multicol}
\usepackage[bookmarks=true]{hyperref}
\usepackage{booktabs}   
\usepackage[dvipsnames]{xcolor}
\usepackage[normalem]{ulem}
\usepackage{amsmath}
\usepackage{amssymb}

\usepackage[caption=false, font=footnotesize]{subfig}

\include{handy_commands}

\include{notation}

\usepackage{graphicx}
\graphicspath{ {images/} }

\title{\LARGE \bf%
Multi-View Picking:\\%
Next-best-view Reaching for Improved Grasping in Clutter%
\vspace{-2mm}
}

\author{Douglas Morrison$^{1}$, Peter Corke$^{1}$ and J\"urgen Leitner$^{1}$%
\vspace{-2mm}
\thanks{This research was supported by the Australian Research Council Centre of Excellence for Robotic Vision (project number CE140100016).}
\thanks{$^{1}$Australian Centre for Robotic Vision (ACRV), Queensland University of Technology (QUT), Brisbane, Australia
}
\thanks{
Contact: {\tt\small doug.morrison@roboticvision.org}}%
}

\begin{document}

\maketitle
\thispagestyle{empty}
\pagestyle{empty}

\begin{abstract}

Camera viewpoint selection is an important aspect of visual grasp detection, especially in clutter where many occlusions are present.  
Where other approaches 
use a static camera position or fixed data collection routines, our Multi-View Picking (MVP) controller uses an active perception approach to choose informative viewpoints based directly on a distribution of grasp pose estimates in real time, reducing uncertainty in the grasp poses caused by clutter and occlusions.
In trials of grasping 20 objects from clutter, our MVP controller achieves 80\% grasp success, outperforming a single-viewpoint grasp detector by 12\%.  We also show that our approach is both more accurate and more efficient than approaches which consider multiple fixed viewpoints.  Code is available at \url{https://github.com/dougsm/mvp_grasp}

\end{abstract}
\vspace{-2mm}

\section{Introduction}

Grasping and transporting objects is a canonical robotics problem which has seen great advancements in recent years, especially with regards to detection of grasp poses for previously unseen objects using only visual information.  As the performance of these visual grasp detection systems has improved, so has the difficulty of standard benchmarking tasks for evaluation, seeing a shift away from grasping relatively simple, isolated objects to grasping geometrically and visually challenging objects in cluttered environments~\cite{mahler2017binpicking}.

The use of cluttered environments and complex objects has resulted in more visually challenging scenarios for grasp detection, with the level of clutter impacting grasp detection performance~\cite{mahler2017binpicking}.  Recent work has shown that
improved visual information from point cloud fusion~\cite{ten2017grasp, arruda2016active} or 
viewpoint selection~\cite{gualtieri2017viewpoint} can improve the quality of grasp estimates in clutter. However, these typically are treated as separate, fixed data gathering step
and do not directly take into account grasp detections from multiple viewpoints.

At the same time, improvements in grasp detection models and computational hardware have seen the time required to visually detect grasps (especially using deep learning methods) decrease from tens of seconds~\cite{lenz2015deep} to less than a second~\cite{mahler2017dex, viereck2017learning} to small fractions of a second~\cite{morrison2018closing}.  Consequently, the largest contributor to grasp execution time has become the movement of the robotic arm executing the grasp, making grasp detection from multiple viewpoints feasible with very little impact to overall execution time.



To address the added difficulties of grasping in clutter, we propose to use the act of reaching towards a grasp as a method of data collection, making it a meaningful part of the grasping pipeline, rather than just a mechanical necessity.

\begin{figure}[t]
  \centering
  \includegraphics[width=0.95\columnwidth]{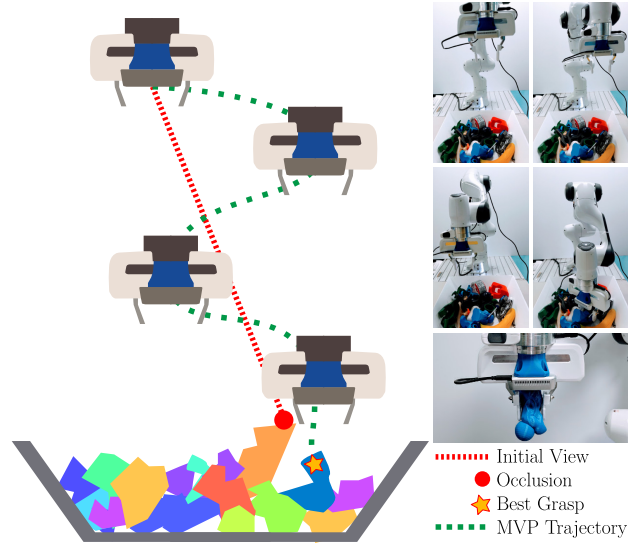}
  \vspace{-1mm}
  \caption{Our Multi-View Picking (MVP) controller considers multiple informative viewpoints for an eye-in-hand camera while reaching for a grasp in clutter to reduce uncertainty in the grasp pose estimate caused by clutter and occlusions.  This allows it to perform high quality grasps that may not be clearly visible from the initial viewpoint.}
  \label{fig:hero}
  \vspace{-4mm}
\end{figure}

To achieve this, we develop the Multi-View Picking (MVP) controller, which selects multiple informative viewpoints for an eye-in-hand camera while reaching to a grasp in order to reduce uncertainty in the grasp pose estimate caused by clutter and occlusion, resulting in an overall improvement in grasp success (\fig{hero}).  Unlike previous works in active perception for grasping which employ object-specific heuristics~\cite{gualtieri2017viewpoint} or a secondary task such as point cloud reconstruction~\cite{arruda2016active, kahn2015active, ten2017grasp}, our approach directly uses entropy in the grasp pose estimation to influence control. 

We validate our approach through trials of grasping from piles of 20 objects in clutter, and compare our results to baselines which represent common visual grasp detection approaches.  Using our MVP controller, we achieve an 80\% success rate in grasping from clutter, a 12\% increase compared to a single-viewpoint grasp detection approach.
Additionally, our method outperforms a baseline which considers multiple viewpoints over a predefined trajectory, achieving a higher grasp success rate with fewer distinct viewpoints and reduced mean time per grasp. This highlights the advantages of using our information-gain based approach which focuses on salient areas, reducing unnecessary data collection.   By varying the cost associated with data collection during reaching, we show that it is possible to trade off between success rate and execution time, allowing the system to be optimised for either raw grasp success rate or overall efficiency.

\section{Related Work}

In order to grasp and transport a wide range of unknown objects, a robotic system cannot rely on using offline information such as object models or object-specific grasp poses.  Instead, it must use geometric information to compute stable grasp poses for previously unseen objects.  Recently, many approaches combining visual inputs with machine-learning techniques -- primarily Convolutional Neural Networks (CNNs) -- have been widely and successfully applied to this problem~\cite{lenz2015deep, morrison2018closing, mahler2017dex, mahler2017binpicking, pinto2016supersizing, ten2017grasp, johns2016deep}, which we refer to as visual grasp detection.  

The robustness of many visual grasp detection systems is sensitive to factors such as sensor noise, robot control inaccuracies and visual occlusion, which is especially prevalent in cluttered environments.  While the detrimental effects of sensor noise and poor robot precision can be minimised as part of the visual grasp detection algorithm, e.g. through sensor noise injection during training~\cite{johns2016deep, mahler2017dex, mahler2017binpicking, viereck2017learning}, overcoming the issue of occlusion in clutter requires multiple camera viewpoints to be considered.  For example, ten \citet{ten2017grasp} showed that computing grasp poses using a fused point cloud from many viewpoints along a predefined trajectory resulted in a 9\% increase in grasp success rate compared to using a point cloud collected from two static cameras.  
Rather than rely on a fixed data collection routine, \citet{arruda2016active} use an active perception approach to choose viewpoints which specifically aid point cloud reconstruction near potential finger contact points in an efficient manner.

Broadly, active perception is defined as the situation where a robot ``adopts strategies for decisions of sensor placement or sensor configuration" in order to perform a task~\cite{chen2011active}.  It is a concept which has been applied to a wide variety of robotic tasks, such as mapping~\cite{stachniss2005information, burgard2005coordinated}, object modelling~\cite{pito1999solution}, object identification~\cite{roy2004active, monica2018contour} and path planning~\cite{velez2011planning}. Common strategies for active perception focus on planning the expected next best action to efficiently minimise measurement uncertainty or maximise information gain via a metric such as Shannon entropy~\cite{vazquez2001viewpoint} or KL divergence~\cite{van2012maximally}.  

Active perception approaches have been applied to robotic grasping in prior work, but rather than directly use grasp quality as a metric rely on a secondary task such as object modelling~\cite{bone2008automated, aleotti2014perception}, point cloud reconstruction while searching for graspable geometry~\cite{arruda2016active, kahn2015active} or localising previously seen objects~\cite{holz2014active} to generate next best view commands.  This is partly due to the difficulty in defining a probability distribution over predicted grasp poses~\cite{gualtieri2017viewpoint}, which we overcome by using a visual grasp detection system which generates a pixelwise distribution of grasp pose estimates~\cite{morrison2018closing}.

\citet{gualtieri2017viewpoint} apply active perception directly to grasp detection.  They compute a distribution of viewpoints for object classes which are likely to improve the quality of detected grasps.  However, this approach requires knowledge of the object's class and doesn't easily generalise to cluttered environments or more complex objects.  They also show improved grasp estimates using a heuristic approach where the camera is aligned to the best detected grasp.



\section{Multi-View Picking}

The choice of camera viewpoints plays an important role in the quality of visual grasp detection.  In this work, we apply active perception techniques to compute the next best viewpoint for a robot with an eye-in-hand camera in real time while reaching for an object.  Unlike previous work, our approach does not rely on any object-specific knowledge or heuristics, does not use a fixed data collection routine and uses visual grasp detection observations directly rather than a secondary metric such as point cloud reconstruction.

\begin{figure*}[tpb]
  \centering
  \includegraphics[width=\textwidth]{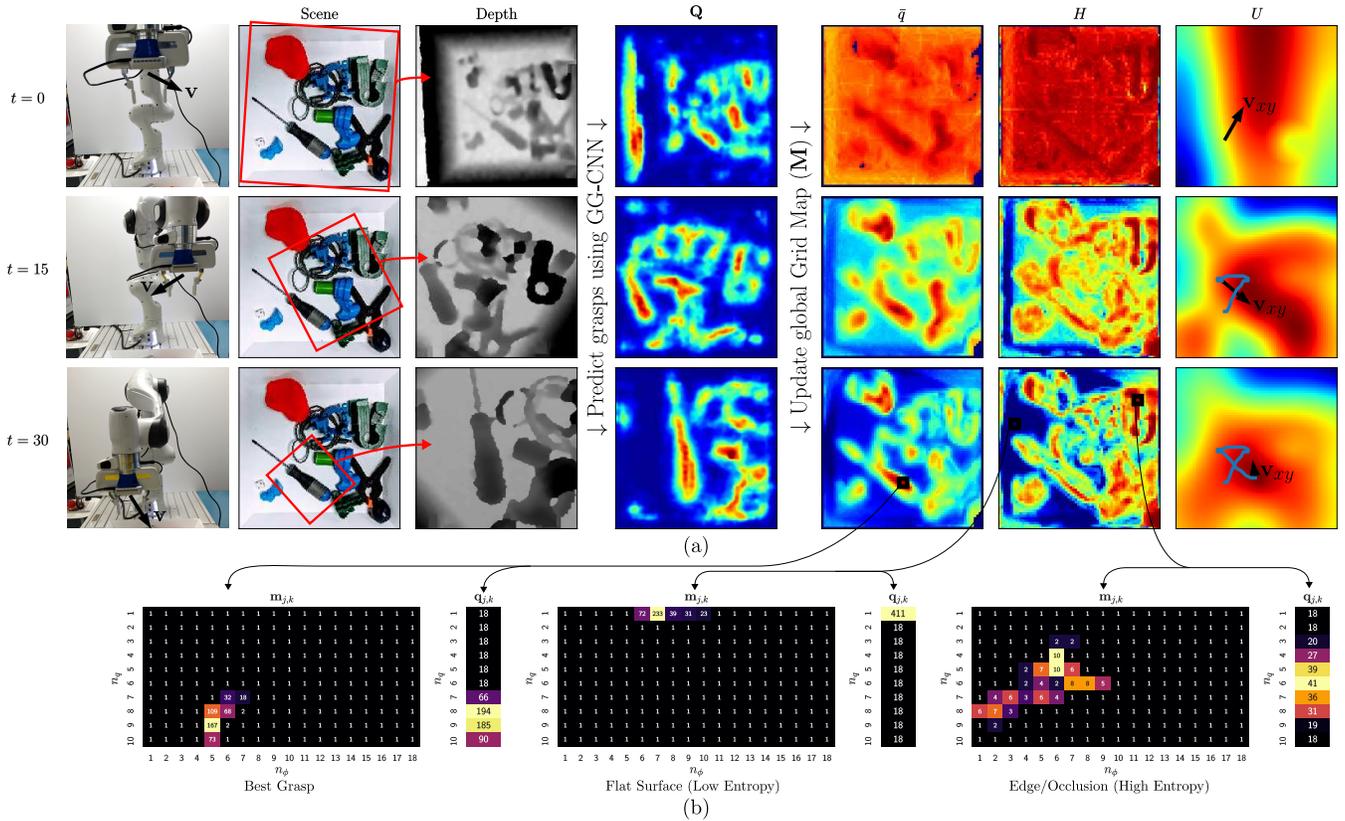}
  \vspace{-7mm}
  \caption{An overview of our MVP controller.  (a) Three time steps during an example grasping trajectory which converges to the point with the best detected grasp pose (the screwdriver handle), considering informative viewpoints corresponding to areas of high expected information gain while reaching. From left to right: The position of the robot (and camera); the collection of objects being grasped; the depth image captured from the current pose $\bp_t$; the quality output of the GG-CNN ($\mathbf{Q}$); the average quality at each grid cell ($\bar{q}$); the entropy at every grid cell ($H$); the utility of a viewpoint ($U$) above each cell in the workspace, from which the velocity command $\bv$ is calculated, and the traversed path superimposed in blue.  (b) The histograms $\bm\ind$ and $\bq\ind$ which represent the counts of observations at, from left to right, grid cells with the best estimated grasp pose, a flat surface with low entropy and, a highly occluded area which results in a high measurement entropy.}
  \label{fig:overview}
  \vspace{-3mm}
\end{figure*}

We develop an information-gain controller, which we call the Multi-View Picking (MVP) controller.  The MVP controller combines visual grasp predictions from multiple camera viewpoints along a trajectory to a grasp pose, and seeks to minimise the uncertainty (entropy) associated with the grasp pose prediction by altering the trajectory to include informative viewpoints, overcoming the challenges associated with visual grasp detection in clutter.  Our implementation of the MVP controller is described in the following sections, with an overview given in \fig{overview}.

\begin{figure}[b]
  \centering
  \includegraphics[width=\columnwidth]{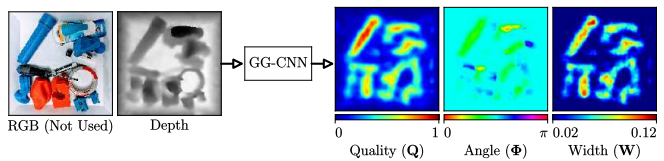}
  \vspace{-5mm}
  \caption{We use the GG-CNN grasp prediction network \cite{morrison2018closing}.  Given a depth image, the GG-CNN predicts the quality, rotation around the vertical axis and gripper width for a grasp at the position corresponding to each pixel.}
  \label{fig:ggcnn}
\end{figure}

\subsection{Viewpoint Trajectory}

We consider the case of a robotic arm with an antipodal gripper and an eye-in-hand camera reaching from an arbitrary starting pose to a grasp pose $\bg$.  At any instance during this motion, the camera's viewpoint $\bp$ is the 3D position of the camera, centred above a point in the workspace $(x,y)$ at a height $z$, which we constrain to be parallel to the xy-plane.  We define a viewpoint trajectory for grasping as a set of $K$ discrete viewpoints $\cT = \{\bp_0, ..., \bp_K\}$ from which we make a visual grasp detection observation while reaching for a grasp.  The initial viewpoint $\bp_0$ is at a fixed height $z_\text{max}$, and the trajectory ends when the camera reaches a height $z_\text{min}$, at which point the best detected grasp is executed.

\subsection{Visual Grasp Detection}

For visual grasp detection, we use the real-time Generative Grasping Convolutional Neural Network (GG-CNN) from \cite{morrison2018closing}.  Given a depth image as input, the GG-CNN produces a pixelwise visual grasp detection $\bx = (\bQ, \bP, \bW)$, representing the grasp quality (the chance of grasp success), angle of rotation around the vertical axis and gripper width at each pixel of the input respectively (\fig{ggcnn}).  Each pixel of $\bx$ represents an antipodal grasp $\bg = (\bc, \phi, w, q)$, parameterised by the grasp's centre position $\bc = (x,y,z)$, rotation $\phi \in [0, \pi]$, gripper width $w$ and quality $q \in [0,1]$.  $\bc$ is computed from the measured depth and the camera's intrinsic parameters.

Other visual grasp detection systems regress a single grasp pose~\cite{kumra2017grasp, redmon2015realtime} or perform classification on sampled grasp candidates~\cite{mahler2017dex, lenz2015deep}, which do not easily lend themselves to defining a probability distribution over grasp estimates.  The GG-CNN is an ideal component of our active perception system because it directly generates a distribution of grasp estimates and also gives real-time computation (approx. 50Hz).  An additional advantage of GG-CNN is that it would be trivial to mask the pixelwise grasp estimates with an off-the-shelf semantic segmentation system if object-specific grasping was required. 

\subsection{Grid Map Representation}

To combine observations at time-steps along the viewpoint trajectory, 
we represent the workspace of the robot as a 2D grid map, $M$, of $J \times K$ cells.  Each cell corresponds an $\gcs \times \gcs$ physical area.  The grid map has the advantage of being computationally efficient compared to other representations, such as Gaussian Processes, which could be used in a similar.
Within each cell $(j,k)$, grasp quality observations ($q$) are counted in a vector $\bq\ind$, discretised into $N_q$ intervals, and combined grasp quality and angle observations ($q$, $\phi$) are counted in a 2-dimensional histogram $\bm\ind$, discretised into $N_q \times N_\phi$ intervals respectively (\fig{overview}b).  These vectors represent the distribution of observations at each point, and form the basis for our information gain approach.  

A grasp at cell $(j,k)$ is parameterised by the mean of observations within that cell:
\begin{equation}
    \bg\ind = (\bc\ind, \bar{\phi}\ind, \bar{w}\ind, \bar{q}\ind)
\end{equation}%
where $\bc\ind$ is the physical position at the centre of the cell and the mean observations are calculated as per below.  (We drop the $j,k$ notation for the sake of readability.)

For a single cell, the mean quality observation $\bar{q}$ is given by:
\begin{equation}
    \bar{q} = \frac{1}{\sum \bq} \sum_{n_q=1}^{N_q} \frac{n_q}{N_q} \bq_{n_q}
\end{equation}%
and the mean angle $\bar{\phi}$ is the vector mean of the angle observations~\cite{mardia2014statistics} weighted by the corresponding grasp quality observations:
\begin{equation}
    \bar{\phi} = \arctan \frac{
    \sum_{n_\phi=1}^{N_\phi} 
    \sum_{n_q=1}^{N_q} 
    \sin(\frac{n_\phi}{N_\phi}\pi) \frac{n_q}{N_q}\bm_{n_\phi,n_q} }
    {\sum_{n_\phi=1}^{N_\phi}  
    \sum_{n_q=1}^{N_q} 
    \cos(\frac{n_\phi}{N_\phi}\pi)\frac{n_q}{N_q}\bm_{n_\phi,n_q}}
\end{equation}

The mean grasp width $\bar{w}$ for a cell is simply the mean of $n$ observations: $\bar{w} = \frac{1}{n} \sum w$

\subsection{MVP Controller}

We formulate our MVP controller using an information gain approach, where 
viewpoints are selected to reduce the uncertainty in the grasp pose observations.  Specifically, we aim to reduce the entropy of grasp quality observations in $M$ which correspond to high quality grasps.  

We can calculate the entropy of the grasp quality observations a single grid cell as:
\begin{equation}
    H(\bq) = - \sum_{n_q = 1}^{N_q} \frac{\bq_{n_q}}{\sum \bq} \log \left( \frac{\bq_{n_q}}{\sum \bq} \right)
\end{equation}
Here, we calculate entropy in the quality observations $\bq$ only, rather than the full distribution of grasp quality and angle in $\bm$ because entropy in the distribution of angle measurements is not always a good indicator of uncertainty in the measurement.  For example a small, spherical object can be graspable from any angle with high quality, so may have a high entropy across $\bm$ but low entropy when considering only $\bq$.

We find that entropy in the distribution of grasp quality measurements is a good candidate for predicting informative viewpoints.  The fifth column of \fig{overview}a shows the entropy of measurements in a semi-cluttered scene at three time steps during a grasp.  As shown in \fig{overview}, areas of high entropy are present around areas of clutter, occlusion and complex geometry where the output of the grasp detector is highly variable and dependent on viewpoint, compared to ``uninteresting" the areas with certain measurements (and hence low entropy) regardless of viewpoint such as the flat bottom surface of the workspace.

Calculating the expected information gain of an observation from a viewpoint, in terms of reduction in entropy, is an intractable problem in this case.  As such, we use the common simplifying assumption that the total entropy of the observed area will provide a good approximation to computing the expected information gain~\cite{thrun2005probabilistic}.  That is, viewpoints which observe areas of high entropy are likely to be more informative (i.e. reduce entropy) than observing areas which already have low entropy.  

Hence, we approximate the expected information gain from an observation at a viewpoint as the weighted sum of entropy of the grid cells observed from that viewpoint:
\begin{equation}
    E[I(M, \bp)] \approx \sum_{j,k \in O_\bp} P(j,k) \cdot H(\bq\ind)
\end{equation}%
where $O_\bp$ is the set of grid cell coordinates observable by the camera from viewpoint $\bp$, and $P(j,k)$ weights points by a Gaussian function based on their distance from the geometric centre of $O$, encouraging the controller to view high-entropy areas front-on rather than peripherally.  

To predict the next best viewpoint, we calculate the utility of moving to a viewpoint centred above each cell in $M$, which represents the desirability of moving to each viewpoint as:

\begin{equation}
    U(M, \bp) = E[I(M, \bp)] - \gamma \cost{\bp}
\end{equation}
where $\cost{\bp}$ is the cost associated with moving to the viewpoint $\bp$, tunable by the \textit{exploration cost} parameter $\gamma$. To encourage the controller to observe areas nearby the best detected grasp, rather than irrelevant distant points, the cost term represents the horizontal ($xy$) distance to the grid cell with the highest average grasp quality, centred at $\bc$:

\begin{equation}
    \cost{\bp} = \left\Vert \bp_{xy} - \bc_{xy} \right\Vert \cdot \left( 1-\frac{\bp_z - z_\text{min}}{z_\text{max}-z_\text{min}} \right)
\end{equation}

The second term scales the cost based on the vertical position of the camera in the trajectory ($\bp_z$).  At the beginning of the trajectory there is zero cost, encouraging exploration of the workspace, which increases linearly as the end-effector descends, causing the controller to converge to the best grasp. 


The MVP controller generates a horizontal velocity command $\bv_{xy}$ in the direction of maximum utility, shown in the final column of \fig{overview}a.  To enable direct comparison between our experiments, we scale the vertical component of velocity $\bv_z$ such that the absolute end-effector velocity $\left\vert\bv\right\vert$ is constant. 

\section{Robotic Experiments}
\label{secn:experiments}

\begin{figure*} 
    \centering
    \hfill
  \subfloat[]{%
       \includegraphics[width=0.35\textwidth]{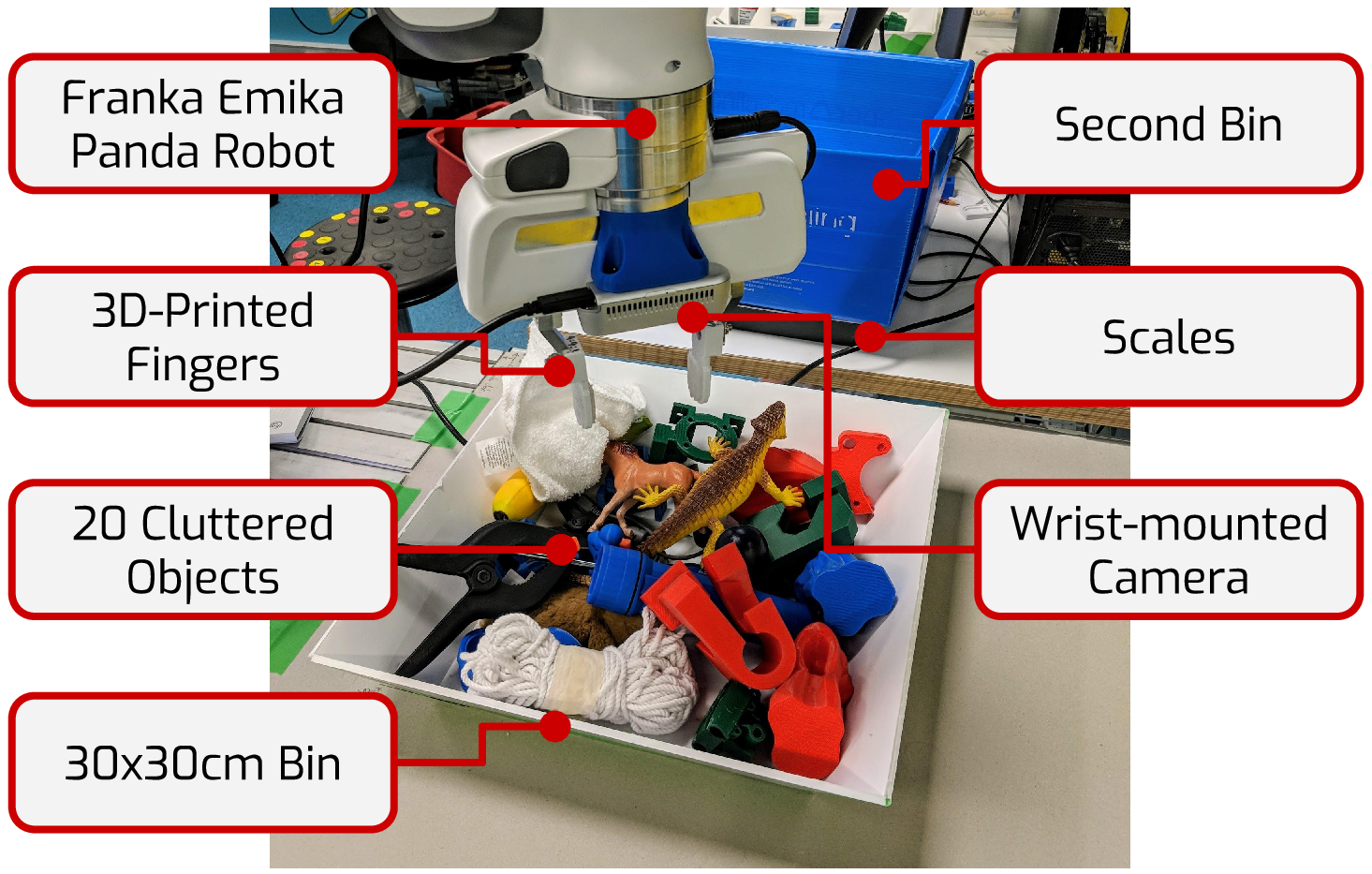}}
    \label{1a}\hfill
  \subfloat[]{%
        \includegraphics[width=0.3\textwidth]{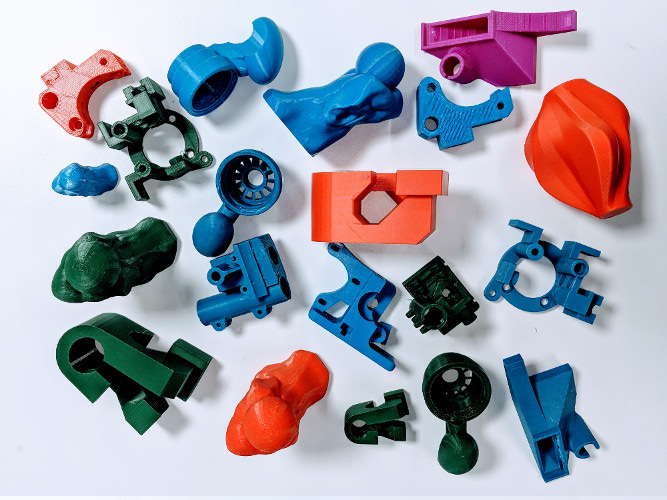}}
    \label{1b}\hfill
  \subfloat[]{%
        \includegraphics[width=0.225\textwidth]{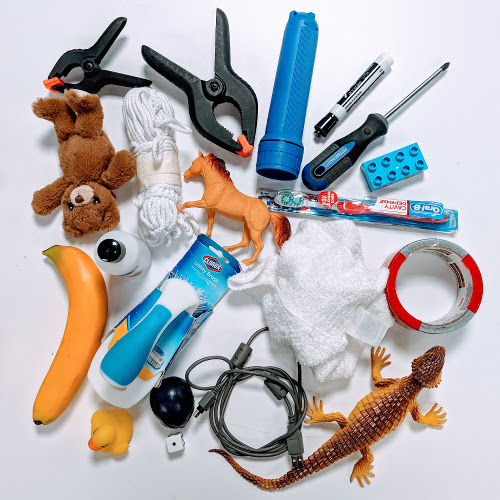}}
    \label{1c}\hfill
  \caption{(a) Our experimental setup. (b) The adversarial object set. (c) The household object set.}
  \label{fig:setup} 
\end{figure*}

We test and validate our approach through a number of trials of robotic grasping in clutter, with the setup is shown in \fig{setup}a.  Our experiments are described in the following sections.  Our software implementation of the system in the form of ROS nodes, primarily written in python, will be made available online. 

\subsection{Equipment}

Experiments are performed using a Franka Emika Panda robot, fitted with custom, 3D-printed gripper fingers using the design from~\cite{guo2017design}.  We use an Intel Realsense D435 depth camera, which is mounted to the robot's end-effector.  

\subsection{Objects}

We use a set of 40 objects, comprising 20 ``adversarial'' 3D-printed objects from the DexNet 2.0~\cite{mahler2017dex} dataset\footnote{3D-printable mesh files are available from\\ \href{https://berkeleyautomation.github.io/dex-net/}{https://berkeleyautomation.github.io/dex-net/}} (\fig{setup}b), and 20 household objects (\fig{setup}c).  The adversarial objects have complex geometry, making them difficult to perceive and grasp.  The household objects contains objects that are a wide variety of sizes and shapes and include deformable objects and visually challenging objects which are transparent or black.

\subsection{Method}

For each experimental run, 20 objects are chosen at random, 10 from each of the adversarial and household sets, and emptied into a $30\times30$cm bin in an unstructured jumble.  The robot then grasps objects one by one and places them into a second bin until all objects have been removed.  A grasp is counted as a success if the object is successfully transported to the second bin.  Scales on the second bin are used to record the success or failure.

We perform 9 experiments using the MVP controller, comprising 7 runs each, varying the \textit{exploration cost} $\gamma$ while keeping all other parameters listed in \tab{exp1_params} constant.  Additionally, we perform baseline experiments for comparison as described in the next section.

\begin{table}[tpb]
    \caption{Parameters used during experiments}
    \label{tab:exp1_params}
    \centering
    \begin{tabular}{@{}llr@{}}
        \toprule
        \multicolumn{2}{l}{ \textbf{Parameter} } & \textbf{Value} \\
        \midrule
        $J$, $K$ & Grid Map Size & 68 \\
        $\gcs$ & Grid Cell Size & 5mm \\
        $N_q$ & Quality Bins & 10 \\
        $N_\phi$ & Angle Bins & 18 \\
        $z_\text{max}$ & Starting Height & 0.5m \\
        $z_\text{min}$ & Final Height & 0.2m \\
        & Controller Update Rate & 10Hz \\
        $\left\vert\bv\right\vert$ & End-effector Velocity (During Reach) & 0.1m/s \\
        $\gamma$ & Exploration Cost & \textbf{Varied} \\

        \bottomrule
        \end{tabular}
\vspace{-3mm}
\end{table}

\subsection{Baselines}
\label{secn:baselines}

We compare our results to three baselines which represent common methods in other robotic grasping work.  We complete 5 runs of each baseline using the method above.  Where relevant, the parameters in \tab{exp1_params} (including the end-effector velocity) are kept constant to allow for the best possible comparison.

\textbf{Single Viewpoint} Most work in visual grasp detection considers only a single viewpoint for grasp detection~\cite{mahler2017dex, mahler2017binpicking, pinto2016supersizing, lenz2015deep, johns2016deep}.  In this baseline we always execute the best grasp detected by the GG-CNN from a single viewpoint centred above the workspace.  

\textbf{Fixed Data Collection} ten \citet{ten2017grasp} increase their grasp success rate by performing point cloud fusion along a fixed trajectory.  Because our method uses grasp estimates rather than point clouds, we use our grid map representation to combine GG-CNN predictions along a fixed, spiral trajectory (\fig{paths}), considering 25 and 50 uniformly spaced viewpoints in two experiments.

\textbf{No Exploration} \citet{gualtieri2017viewpoint} showed an increase in grasp success by aligning the camera to the axis of the best detected grasp.  In this baseline we disable the exploration of our MVP controller, instead always generating a velocity command to align the camera to the best detected grasp at each time step.

\section{Results}
\begin{table*}[t]
    \caption{Experimental results for grasping objects in clutter, using our MVP controller and the three baseline cases.}
    \label{tab:results}
    \centering    
    \begin{tabular}{@{}llllllllllcllll@{}}
        \toprule
        & \multicolumn{9}{c}{ \textbf{Ours (MVP Controller)} } && \multicolumn{4}{c}{ \textbf{Baselines} } \\
        \cmidrule{2-10} \cmidrule{12-15}
        & \multicolumn{9}{c}{ \textit{Exploration Cost} ($\gamma$) } && \textit{Single View} &  \textit{No Expl.} & \multicolumn{2}{c}{ \textit{Fixed Data Collection} } \\
        \cmidrule{14-15}
        & 0.0 & 0.05 & 0.1 & 0.2 & 0.3 & 0.4 & 0.5 & 0.6 & 0.7 & & \textit{} & \textit{} & 25 Views & 50 Views \\
        \cmidrule{2-10} \cmidrule{12-15}
        Total Attempts & 174 & 175 & 178 & 171 & 181 & 182 & 179 & 189 & 188 && 196 & 187 & 137 & 125 \\
        Failures & 35 & 35 & 38 & 36 & 42 & 42 & 45 & 50 & 49 && 62 & 48 & 37 & 28 \\
        Mean Viewpoints & 44 & 43 & 39 & 36 & 35 & 35 & 35 & 35 & 35 && 1 & 35 & 25 & 50 \\
        \cmidrule{2-10} \cmidrule{12-15}
        Success Rate (\%) & \textbf{80} & \textbf{80} & 79 & 79 & 77 & 77 & 75 & 74 & 74 && 68 & 74 & 73 & 78 \\
        Mean Time (s) & 10.5 & 10.2 & 9.8 & 9.2 & 9.0 & 9.1 & 9.0 & 9.1 & 9.1 && \textbf{8.8} & 9.1 & 11.4 & 11.4 \\
        MPPH & 273 & 282 & 288 & \textbf{308} & 307 & 305 & 299 & 292 & 292 && 281 & 293 & 230 & 245 \\
        \bottomrule
        \end{tabular}
\vspace{-2mm}
\end{table*}

The results of our experiments are shown in \tab{results}.  We assess each experiment based on three metrics:
\begin{itemize}
    \item \textit{Success Rate:} The overall ratio of successful grasps to grasp attempts across all runs.
    \item \textit{Mean Time per Pick:} The average time per grasp attempt (in seconds), regardless of success.
    \item \textit{Mean Picks Per Hour (MPPH):} The overall efficiency of the system representing the average rate of successful picks per hour calculated as $\text{3600 [s/h]} \div (\text{Mean Time per Pick [s]} ) \times (\text{Success Rate})$
\end{itemize}

It is important to note that we include the MPPH measurement as a way of comparing the different results, so are primarily concerned with the relative difference of values rather than their overall magnitude which is highly dependent on the fixed end-effector velocity that we use.

\subsection{Exploration During Grasping}

\begin{figure}[tpb]
  \centering
  \includegraphics[width=\columnwidth]{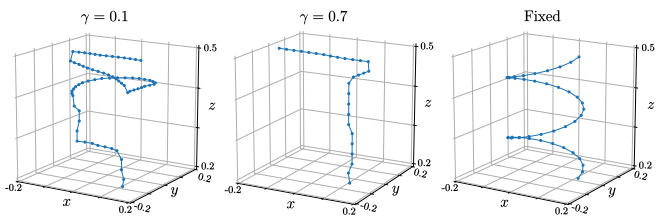}
  \vspace{-5mm}
  \caption{Three viewpoint trajectories, for $\gamma=0.1$, $\gamma=0.7$ and the \textit{Fixed Data Collection} baseline.}
  \label{fig:paths}
  \vspace{-4mm}
\end{figure}

We first investigate the effect of trading off between exploration and execution time.  Varying the \textit{exploration cost} from $\gamma=0.7$ (minimum exploration) to $\gamma=0.0$ (maximum exploration) results in a 6\% increase in grasp success rate, improving from 74\% to 80\%, at the expense of 1.4 seconds per grasp on average.  Two example trajectories for $\gamma=0.1$ and $\gamma=0.7$ are shown in \fig{paths}.  

As the increased time associated with exploration does not scale linearly with the increase in grasp success rate, the overall efficiency of the system, measured in MPPH, is maximised between the two extremes, for $\gamma=0.2$, where there is an increase in grasp success rate (79\%) but minimal extra time cost.  As a result, the MVP controller can be optimised for either raw grasp success rate or overall efficiency by adjusting the cost of exploration. 

As $\gamma$ is increased, the results begin to plateau in the range 0.5 to 0.7 due to the the cost term becoming dominant in this range and causing to controller to perform minimal exploration and converge directly to the grasp. 

\subsection{Comparison to Baselines}

\textbf{Single Viewpoint}  Our approach outperforms the single viewpoint baseline approach in terms of both success rate, improving up to 12\% (68\% vs. 80\%), and MPPH, increasing by approximately 10\% from 281 to 308 despite an increased mean time per pick.  This reinforces the assertion that considering multiple viewpoints is an effective way to overcome the visual challenges associated with grasping in clutter. 

\textbf{No Exploration} The \textit{No Exploration} baseline achieves similar results to our MVP controller when using a high \textit{exploration cost}, which is unsurprising since the high \textit{exploration cost} results in minimal exploration.  However, for lower values of $\gamma$, our MVP controller outperforms this baseline by 6\% success rate, highlighting the added benefit of actively exploring based on uncertainty compared to using a heuristic such as aligning with the best detected grasp.

\textbf{Fixed Data Collection}  The \textit{Fixed Data Collection} baseline reinforces the idea that incorporating multiple viewpoints can improve the success rate of a grasping system, with both experiments outperforming the \textit{Single Viewpoint} baseline.  Furthermore, the 50-viewpoint experiment outperforms the 25-viewpoint experiment by 5\% grasp success rate.  However, because the fixed trajectory always views the entire workspace uniformly, it results in a constant execution time which is longer than all other experiments and is unable to focus on salient areas of the workspace.  As a result, our MVP controller outperforms this baseline with regards to all metrics, including a higher grasp success rate with fewer viewpoints.  The main advantage comes from our information gain approach, which is able to focus on salient areas of the workspace to reduce unnecessary data collection.


\subsection{Automatically Adapting to Scene Complexity}

\begin{figure}[tpb]
  \centering
  \includegraphics[width=0.97\columnwidth]{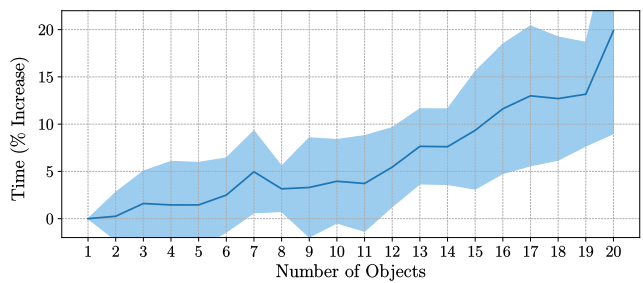}
  \vspace{-2mm}
  \caption{Mean time per pick increases with the number of objects present, compared to the mean time for one object, averaged over all experimental runs (all values of $\gamma$). Shaded area shows one standard deviation.}
  \label{fig:timevsobjects}
  \vspace{-4mm}
\end{figure}

As shown with the \textit{Fixed Data Collection} baseline, using a set of fixed viewpoints for data collection can increase grasp detection accuracy.  However, it results in longer, constant execution times, and is unable to focus attention on ``interesting" parts of the scene.  In contrast, our MVP controller is able to actively adapt to the complexity of the scene and provides more a more efficient data collection process.  \fig{timevsobjects} shows that for our MVP controller the mean time per pick is dependent on the number of objects in the workspace.  As the number of objects in the scene increases, and with it the amount of clutter and potential occlusions, so does the mean time per pick, taking on average 20\% longer when 20 objects are present compared to a single object. 


\section{Conclusion}

We presented an active perception approach to grasping in clutter, in which we consider visual grasp detections from multiple viewpoints while reaching.  Our work reinforces the importance of viewpoint selection and combining data from multiple viewpoints when grasping in clutter, but in contrast to previous work our Multi-View Picking controller uses an information-gain approach to select informative viewpoints that directly seek to reduce entropy in the grasp pose estimates caused by clutter and occlusions.  

We validate our approach with several experiments in grasping from clutter.  Our MVP controller achieves up to 80\% grasp success while picking from cluttered piles of up to 20 objects, including adversarial objects with complex geometry, outperforming a single-viewpoint method by 12\%.  

Additionally, by using an information-gain approach, our MVP controller is able to adapt to the complexity of the scene, unlike other approaches which rely on fixed data collection routines as part of a visual grasp detection pipeline.  Compared to such a method, our approach results in a higher grasp success rate while also being more efficient, requiring fewer viewpoints and less time per grasp. 






\printbibliography

\end{document}

%% file: handy_commands.tex
\newcommand{\fig}[1]  {Fig.~\ref{fig:#1}}		
\newcommand{\tab}[1]  {Table~\ref{tab:#1}}		



%% file: notation.tex
\newcommand{\cT}{\mathcal{T}}  
\newcommand{\bp}{\mathbf{p}}   

\newcommand{\bg}{\mathbf{g}}  
\newcommand{\bc}{\mathbf{c}}  
\newcommand{\bQ}{\mathbf{Q}}  
\newcommand{\bP}{\mathbf{\Phi}}  
\newcommand{\bW}{\mathbf{W}}  

\newcommand{\bq}{\mathbf{q}}  
\newcommand{\bm}{\mathbf{m}}  
\newcommand{\bv}{\mathbf{v}}  
\newcommand{\bx}{\mathbf{x}}  

\newcommand{\gcs}{u}  
\newcommand{\cost}[1]{r(#1)}  
\newcommand{\ind}{_{j,k}}